\documentclass[journal]{IEEEtran}
\ifCLASSINFOpdf \else \fi

\usepackage{changes}

\usepackage{caption}
\usepackage{graphicx}
\graphicspath{{./figs/}}
\usepackage{epsfig} 
\usepackage{adjustbox}
\usepackage{siunitx}
\usepackage{booktabs}
\usepackage{multicol}  
\usepackage{multirow}  
\usepackage{xcolor}
\usepackage{soul}

\usepackage{amsmath,amssymb}
\usepackage{graphicx,graphics,color,psfrag}
\usepackage{cite,balance}
\usepackage{caption}
\usepackage{accents}
\usepackage{bm}
\usepackage[english]{babel}
\usepackage{multirow}
\usepackage{enumerate}
\usepackage{cases}
\usepackage{stfloats}
\usepackage{dsfont}
\usepackage{color,soul}
\usepackage{amsfonts}
\usepackage{cite,graphicx,amsmath,amssymb}
\usepackage{fancyhdr}
\usepackage{hhline}
\usepackage{graphicx}
\usepackage{array,color}
\usepackage{booktabs}
\usepackage{diagbox}
\usepackage{indentfirst}
\usepackage{url}
\usepackage{amsthm}

\usepackage{amsmath}

\usepackage{lipsum}

  {
      \theoremstyle{plain}

  }
\usepackage{algorithm}
\usepackage{algorithmic}
\usepackage{array}
\ifCLASSOPTIONcompsoc
 \usepackage[caption=false,font=normalsize,labelfont=sf,textfont=sf]{subfig}
\else
 \usepackage[caption=false,font=footnotesize]{subfig}
\fi

\usepackage{url}
\usepackage{fancyhdr}
\usepackage{color}
\usepackage{breqn}
\usepackage{stfloats}
\newcommand{\Rmnum}[1]{\expandafter\@slowromancap\romannumeral #1@}

\theoremstyle{definition}

\newtheorem{rem}{Remark}

\begin{document}

\title{RadioGen3D: 3D Radio Map Generation via Adversarial Learning on Large-Scale Synthetic Data}

\author{Junshen Chen, Angzi Xu, Zezhong Zhang,~\IEEEmembership{Member, IEEE}, Shiyao Zhang,~\IEEEmembership{Member, IEEE} \\Junting Chen,~\IEEEmembership{Member, IEEE}, and Shuguang Cui,~\IEEEmembership{Fellow, IEEE}
\thanks{J. Chen, A. Xu, Z. Zhang, J. Chen and S. Cui are with the School of Science and Engineering (SSE), the Shenzhen Future Network of Intelligence Institute (FNii-Shenzhen), and the Guangdong Provincial Key Laboratory of Future Networks of Intelligence, the Chinese University of Hong Kong at Shenzhen, China, 518066 (Email: \{225015018, 224010109\}@link.cuhk.edu.cn, \{zhangzezhong, chenjunting, shuguangcui\}@cuhk.edu.cn). S. Zhang is with the School of Advanced Engineering, Great Bay University, Dongguan, China, 523000 (Email: zhangshiyao@gbu.edu.cn).}
}



\maketitle


\begin{abstract}
Radio maps serve as a fundamental basis for efficient radio resource management, enabling diverse applications in future 6G and low-altitude networks, such as network resource optimization and drone trajectory planning. 
In recent years, deep learning (DL) techniques have been widely adopted for radio map estimation (RME), significantly outperforming conventional ray-tracing methods in efficiency while maintaining comparable accuracy.
However, most existing DL-based RME approaches are confined to 2D near-ground scenarios and fail to capture essential 3D signal propagation characteristics. This limitation stems primarily from the scarcity of 3D radio map data and the difficulties in training 3D models. Moreover, the effect of antenna polarization is usually missing in existing solutions. 
To address these limitations, this paper presents the RadioGen3D framework. It first proposes a novel data synthesis method capable of generating high-quality 3D radio map data with significantly higher efficiency and flexibility than traditional ray-tracing methods. Specifically, a parametric target model is established that captures both 2D ray-tracing and 3D channel fading characteristics. Multiple realistic coefficient combinations are derived from minimal real measurements to cover diverse signal propagation characteristics, enabling the construction of a large-scale synthetic dataset \emph{Radio3DMix}.
Building upon the synthetic dataset, we further propose a 3D model training scheme based on a conditional generative adversarial network (cGAN), which yields a 3D U-Net capable of accurate RME with different input feature combinations. 
Experimental results demonstrate that RadioGen3D surpasses all baselines in both estimation accuracy and speed, owing to its data sufficiency and 3D model structure. Its generalization capability is also verified by the successful knowledge transfer in the fine-tuning experiment, confirming the superiority of the proposed framework.
\end{abstract}

\begin{IEEEkeywords}
3D radio map estimation, deep learning, received signal strength, conditional GAN, ray-tracing
\end{IEEEkeywords}

\IEEEpeerreviewmaketitle

\section{Introduction}
The rapid development of sixth-generation (6G) communications has led to the emergence of a growing number of intelligent applications, such as autonomous drone swarms and remote collaborative robotics, imposing stringent requirements on data transmission with high rates and low latency \cite{shi2023task,wang2023road}. This has created a critical need for adaptive radio resource management in current resource-constrained network to support the enormous communication-sensitive applications \cite{romero2022radio,feng2025recent,management1}. To achieve this, radio map estimation (RME) serves as a crucial technique for characterizing the radio resource distributions, which focuses on key features such as received signal strength (RSS), path gain, or even the interference strength \cite{romero2022radio,liu2023uav,levie2021radiounet}. Consequently, the concept of radio map, also referred to as the channel knowledge map \cite{CKM1,CKM2,CKM3}, has drawn increasing attention recently, and some RME solutions have been proposed to enable comprehensive sensing of the surrounding radio environment.

Conventional RME methods are generally categorized into two classes, i.e. data-driven and model-driven. Data-driven RME methods includes Kriging \cite{van2004kriging}, tensor completion \cite{shrestha2022deep,tensorJT}, and kernel-based learning \cite{romero2022radio}, which mainly rely on interpolation or regression by exploiting sparse measurements. For instance, a least-squares problem is formulated in \cite{liu2023uav} to jointly estimate virtual obstacle maps and propagation parameters via regression. In \cite{chen2025dynamic}, the authors propose efficient reconstruction of multi-dimensional spatial-spectral-temporal radio maps via tensor completion, which exploits the multi-domain low-rank structures of sparse measurements \cite{chen2025dynamic}. However, due to the challenge of incorporating environment information, data-driven methods usually suffer from limited accuracy. 
In model-driven methods, the most representative one is ray-tracing, which simulates the physical propagation of signals including reflection, diffraction, and scattering, providing accurate predictions on key channel characteristics such as RSS and multi-path effects \cite{deschamps2005ray,rizk1997two}. Despite being an environment-aware RME method, ray-tracing  suffers from high computational complexity. The simulation process usually requires substantial computation time, ranging from minutes to hours depending on the scenario, limiting its application in latency-sensitive tasks. Therefore, both data-driven and model-driven methods cannot achieve high accuracy and computational efficiency simultaneously in complex, large-scale environments.

In recent years, with the rapid advancement of artificial intelligence, deep learning (DL) techniques have attracted significant attention due to their capabilities in feature representation learning, which align closely with the requirements of RME. Moreover, the low computational complexity of deep neural networks (DNNs) enables highly efficient inference, supporting their use in latency-sensitive tasks. Some pioneering studies have demonstrated the effectiveness of DL-based solutions, achieving near-real-time inference speed and estimation accuracy comparable to ray-tracing methods by leveraging simulated datasets generated through ray-tracing \cite{levie2021radiounet,teganya2021deep,zhang2023rme,zhang2024fast,shibli2024data,ronneberger2015u,chen2023graph,wang2024radiodiff}. 
Specifically, DL-based environment-aware RME is first brought into reality by \cite{levie2021radiounet}, where a U-Net model trained on a ray-tracing simulated dataset is employed for RME based on the transmitter and environment information. Recently, diverse learning paradigms and model structure are further explored, providing significant improvements in both learning efficiency and robustness. 
For example, RME-GAN proposed in \cite{zhang2023rme} leverage generative adversarial training to capture both global propagation and local shadowing effects from sparse measurements, enabling high estimation accuracy with non-uniform sampling. In \cite{zhang2024fast}, conditional GAN is also adopted, providing a solution capable of accurate estimation that does not require explicit transmitter or perfect environment information. Moreover, a graph neural network (GNN) is used in \cite{shibli2024data} to model spatial dependencies directly on a graph representation of the geographic area, allowing effective utilization of sparse measurements to estimate radio environment maps by capturing non-Euclidean spatial correlations.
Recently, the authors in \cite{wang2025radiodiff} propose RadioDiff-3D, where diffusion models are adopted to train a 3D U-Net by leveraging their iterative denoising and attention mechanisms for high-quality RME. In solutions leveraging generative techniques, such as GAN and diffusion models, RME is interpreted as  an image generation task, thus naturally termed \emph{radio map generation}.

However, most existing DL-based methods focus on near-ground 2D RME and thus overlook essential 3D signal propagation effects, such as channel fading along the height dimension and antenna polarization. This limitation restricts their application to 3D RME, which is essential in low-altitude networks as the vertical distribution of signal strength plays a critical role. 
In fact, this issue stems primarily from the scarcity of 3D radio map data as well as the difficulties in training 3D models.
Firstly, acquiring 3D radio map data is inherently difficult. Real-world data collection requires complex aerial sensing systems (e.g., UAV-based platforms), and the process is susceptible to interference, noise, and systematic bias, resulting in limited high-quality samples. 
While ray-tracing methods offer an alternative via simulation, generating 3D maps remains computationally expensive, even with advanced tools like the Sionna toolbox \cite{aoudia2025sionna} recently launched by NVIDIA. Though an exemplary 3D radio map dataset is provided in our previous work \cite{wang2025radiodiff}, it is limited to single-transmitter scenarios with omnidirectional antennas. Thus, building a comprehensive 3D dataset remains a major challenge. Secondly, designing and training 3D neural networks presents additional difficulties. Intuitively, the complex nature of 3D propagation, coupled with the high parameter count of 3D models, often leads to unstable training and low accuracy if large-scale, high-quality datasets are unavailable, not to mention the substantial memory and computational costs during training. Notably, even when 3D data is used, the solution proposed in \cite{wang2025radiodiff} still rely on dense sampling and results in an inferior estimation accuracy compared to 2D estimation. Moreover, to circumvent the challenges of developing such a 3D model, recent studies have explored alternative approaches using 3D neural radiance fields (3D-NeRF) or 3D Gaussian splatting (3DGS) \cite{mildenhall2021nerf,wen2025wrf,wen2026bridging}. However, these methods are inherently incompatible with multi-modal inputs like sparse measurements and rely on computationally intensive point-by-point inference, resulting in a time-consuming rendering process. As a result, both the shortage of 3D radio map data and the absence of dedicated 3D model architectures continue to hinder the progress in environmental-aware 3D RME.

To address the above limitations, we proposed the framework of RadioGen3D in this paper, comprising an efficient data synthesis method and a 3D model training scheme. Specifically, we first establish a parametric target model that captures both 2D ray-tracing and 3D propagation effects, including antenna polarization. The model coefficients are derived from sparse real measurements collected by a self-developed UAV platform \cite{zhonghao}. Multiple feasible coefficient combinations are then derived to represent diverse propagation characteristics, enabling the construction of a comprehensive large-scale synthetic dataset \emph{Radio3DMix}. Leveraging this dataset, we further propose a model training scheme based on a conditional generative adversarial network (cGAN), yielding a 3D U-Net with high estimation accuracy and strong generalization capability. The main contributions of this paper are summarized as follows.

\begin{itemize}
    \item \textbf{Efficient Synthesis of 3D Radio Map Data:} To tackle the issue of data scarcity in 3D RME, we propose a novel method for efficient and flexible generation of high-quality 3D radio map data. A parametric target model, composed of a pre-trained 2D U-Net and empirical correction terms, is first established. Its coefficients are determined via least-squares regression based on a few sets of real measurements. By exploiting multiple feasible coefficient combinations, we generate a large amount of data that covers a wide range of realistic signal propagation characteristics, which are compiled into a large-scale dataset Radio3DMix. With controllable coefficients and linear computational complexity, the generation process achieves both high efficiency and notable flexibility. The generated data also exhibits physically consistent 3D propagation effects, including path loss, shadowing, and antenna polarization. The effectiveness of Radio3DMix is strongly demonstrated through model training and fine-tuning experiments, leading to high estimation accuracy and robust generalization.
    \item \textbf{Fast and Accurate 3D RME Enabled by cGAN:} Based on the large-scale synthetic 3D dataset Radio3DMix, we further propose a cGAN-based model training scheme, where the model capability in radio map generation is effectively enhanced through an adversarial training process.  Given a conditional input, a cGAN establishes a deterministic mapping from the input to the output radio map. This yields higher accuracy compared to other generative methods such as diffusion models, which may produce varying results due to different noise inputs. Moreover, a 3D U-Net is selected as the generator to exploit the structural compatibility with the pre-trained 2D U-Net used in the dataset construction process. Taking advantage of the data sufficiency, adversarial training, and model alignment, the proposed method achieves superior estimation accuracy, outperforming all existing baselines. The cGAN architecture also entails lower computational complexity than diffusion-based or large-model approaches. 
    \item \textbf{Strong Model Generalization Capability:} In addition to performance evaluation on the synthetic Radio3DMix dataset, we conduct a fine-tuning experiment to adapt the pre-trained 3D U-Net to a small-scale 3D ray-tracing dataset \emph{Radio3D-RT}. This dataset exhibits signal propagation characteristics distinct from Radio3DMix while sharing high similarities with the ray-tracing datasets commonly used in relevant existing works. Experimental results show that the fine-tuned 3D U-Net achieves higher estimation accuracy than baseline methods, validating successful knowledge transfer and demonstrating the effectiveness and superiority of the proposed RadioGen3D framework, including both the Radio3DMix dataset and the cGAN-based training scheme.
\end{itemize}

\section{System Model}\label{system-model}
In this section, we first present the communication model and define the radio map as a function of the geographical environment, the transmitter information, and sparse measurements. The learning problem is formulated subsequently.
\subsection{Communication Model}\label{communication-model}
We consider a three-dimensional (3D) rectangular region $\mathcal{R}$ in space with dimensions $W$ (width), $D$ (depth), and $H$ (height) in the presence of $T$ transmitters and $J$ users, denoted as $\mathcal{T}$ and $\mathcal{J}$, respectively. The users, considered as receivers, are able to measure and record the instantaneous received signal strength (RSS). The RSS measurements (or their statistics) at each user are then uploaded to the server in the next uplink time slot, along with the user location. Given negligible communication overhead for uploading such information, we assume the server can access it in real time. According to the physical model of signal propagation, the received signal at the $j$-th receiver can be expressed as
\begin{align}\label{1}
    Y_j = \sum_{i\in \mathcal{T}}\sqrt{P_i\cdot\rho_{i,j}}h_{i,j}X_i + N_j,
\end{align}
where $P_i$ is the transmit power, $X_i \sim \mathcal{CN} (0, 1)$ is the transmitted symbol, and $N_j \sim \mathcal{CN} (0, \sigma_j^2)$ is the Gaussian noise at the receiver. The large-scale fading coefficient of the channel between transmitter $i$ and user $j$ is  $\rho_{i,j}=\rho_{i,j}^{\mathrm{sh}} \cdot \rho_{i,j}^{\mathrm{path}}$, which is composed of the shadowing effect $\rho_{i,j}^{\mathrm{sh}}$ and pathloss $\rho_{i,j}^{\mathrm{path}}$. The corresponding small-scale fading coefficient is denoted as $h_{i,j}\sim\mathcal{CN}(0, 1)$, respectively. According to \eqref{1}, the average RSS in time can be represented as
\begin{align}\label{2}
    \mathbb{E}[\left\|Y_j \right\|^2] = \sum_{i\in \mathcal{T}}P_i\cdot \rho_{i,j} + \sigma_j^2,
\end{align}
which characterizes the long-term communication quality at user $j$. As large-scale fading $\rho_{i,j}$ stems from the interaction between the propagated signals and the geographical environment, and the average noise power $\sigma_j^2$ at the receiver is usually time-invariant, then the aggregation of the average RSS across every location within the region $\mathcal{R}$ can be compactly represented as a radio map $\mathbf{P}\in \mathbb{R}^{W\times D\times H}$, given as
\begin{align}\label{model1}
    \mathbf{P} &= F_1\left(\mathbf{T}, \mathbf{E} \right) = f_1\left(\mathbf{T}, \mathbf{E} \right) + \Sigma,
\end{align}
where the tensor $\mathbf{T}\in \mathbb{R}^{W\times D\times H}$ denotes the transmitter locations and transmit powers by assigning $\mathbf{T}_{\mathbf{q}_i} \triangleq \mathbf{T}_{(x_i,y_i,z_i)} =P_i$ for transmitter $i$ located at $\mathbf{q}_i = [x_i,y_i,z_i]^T$, and setting all other elements to $0$. The tensor $\mathbf{E}\in \{0,1\}^{W\times D\times H}$ maps the geographical environment to a 3D occupancy grid, with $1$ indicating occupied space and $0$ indicating free space. The tensor $\boldsymbol{\Sigma}\in \mathbb{R}^{W\times D\times H}$ is the corresponding noise power map, where $\boldsymbol{\Sigma}_{\mathbf{u}_j} \triangleq \boldsymbol{\Sigma}_{(x_j,y_j,z_j)} = \sigma_j^2$, given the location of user $j$ to be $\mathbf{u}_j = [x_j,y_j,z_j]^T$. 

The above equation maps the transmitter and environment information to the radio map with a function $F_1(\cdot)$, which coincides with the principle of ray-tracing. However, in real-world wireless networks, it is difficult to guarantee access to the transmitter information $\mathbf{T}$ in real-time. Therefore, we also provide another viable approach which substitutes the transmitter information $\mathbf{T}$ with sparse RSS measurements across the distributed users $\mathcal{J}$, represented by a tensor $\mathbf{S}\in \mathbb{R}^{W\times D\times H}$. The tensor of sparse measurements is defined as $\mathbf{S}_{\mathbf{u}_j}=\mathbf{P}_{\mathbf{u}_j}$ if $j\in \mathcal{K}$, and $\mathbf{S}_{\mathbf{u}_j}=0$ otherwise. The sparsity of transmitters implies a one-to-one mapping between the transmitter information and the sparse RSS measurements. Hence, the transmitter information can be represented as
\begin{align}\label{5}
    \mathbf{T} = h(\mathbf{S}, \mathbf{E} ).
\end{align}
Based on \eqref{model1} and \eqref{5}, the radio map is further modeled as
\begin{align}\label{model2}
    \mathbf{P} = F_2\left(\mathbf{S}, \mathbf{E} \right) = F_1\left(h(\mathbf{S}, \mathbf{E} ), \mathbf{E} \right),
\end{align}
which is a function of the sparse RSS measurements $\mathbf{S}$ and environment information $\mathbf{E}$. 

However, both the expressions of $F_1(\cdot)$ and $F_2(\cdot)$ are highly intractable. To address this issue, we translate RME as a data-and-model-driven learning problem, where a DNN is trained to approximate functions $F_1(\cdot)$ or $F_2(\cdot)$ with high fidelity.
Moreover, as an extension, it is also possible to exploit both the transmitter information and the sparse measurements to enhance the inference accuracy of the DNN model, thereby obtaining enhanced variants of the models in \eqref{model1} and \eqref{model2}. In this case, the radio map is modeled as 
\begin{align}\label{model3}
    \mathbf{P} = F_3\left( \mathbf{T}, \mathbf{S}, \mathbf{E} \right).
\end{align}
With the three types of radio map modeling, we formulate the learning problem in the following.

\subsection{Problem Formulation}\label{sec-problem-formulation}
Suppose we have a dataset $\mathcal{D}$ where each data sample is composed of a \emph{complete feature vector} $\mathbf{X}_{\mathsf{full}} \triangleq [\mathbf{T}, \mathbf{S}, \mathbf{E}]$ and a corresponding label $\mathbf{P}$ representing the ground-truth radio map. 
For each data sample, depending on the three types of radio map modeling in \eqref{model1}, \eqref{model2}, and \eqref{model3}, we can construct three corresponding feature vectors from $\mathbf{X}_{\mathsf{full}}$, i.e., $\mathbf{X} = [\mathbf{T}, \mathbf{E}]$, $\mathbf{X} = [\mathbf{S}, \mathbf{E}]$, or $\mathbf{X} = [\mathbf{T}, \mathbf{S}, \mathbf{E}]$, respectively.
In this unified formulation, the learning problem is expressed as
\begin{align}\label{problem-formulation}
     \min_{\Theta_{G}} \sum_{(\mathbf{X}, \mathbf{P})\in \mathcal{D}} \mathrm{Dist}\left(G(\mathbf{X}; \boldsymbol{\Theta}_{G}), \mathbf{P} \right),
\end{align}
regardless of the specific radio map modeling type. In the above equation, the function $G(\cdot;\Theta_{G})$ denotes the DNN model parameterized by $\boldsymbol{\Theta}_G$. The function $\mathrm{Dist}(\cdot)$ is the error function that quantifies the discrepancy between the prediction $G(\mathbf{X}; \boldsymbol{\Theta}_{G})$ and the label $\mathbf{P}$.

Although the formulation in \eqref{problem-formulation} is common in deep learning, and there exist some DL-based solutions to 2D radio map estimation, solving the problem via training still faces the following two challenges. Firstly, training a 3D model is usually much more difficult than the 2D training case since the input and output captures more complicated features, e.g., 3D environments, 3D channel fading, and antenna polarization effects, which are usually missing in 2D RME tasks. This raises the demand for both appropriate model selection and a large amount of high-quality data. Secondly, it is challenging to acquire ground-truth radio map labels for every data sample in a dataset. 
Though some existing ray-tracing simulated datasets are available, e.g., RadioMapSeer\cite{levie2021radiounet}, RadioGAT\cite{li2024radiogat}, and UrbanRadio3D\cite{wang2025radiodiff}, they are limited to specific scenarios. Moreover, generating new data tailored to customized tasks via ray-tracing still remains a time-consuming process. 
Therefore, to address the above issues, we propose RadioGen3D, a novel framework for efficient dataset generation and 3D model training, which is elaborated in the following sections.

\section{Efficient Data Synthesis and cGAN-based Radio Map Generation}\label{sec:RadioGen3D}
\begin{figure*}[t]
    \centering
    \includegraphics[width=0.75\linewidth]
    {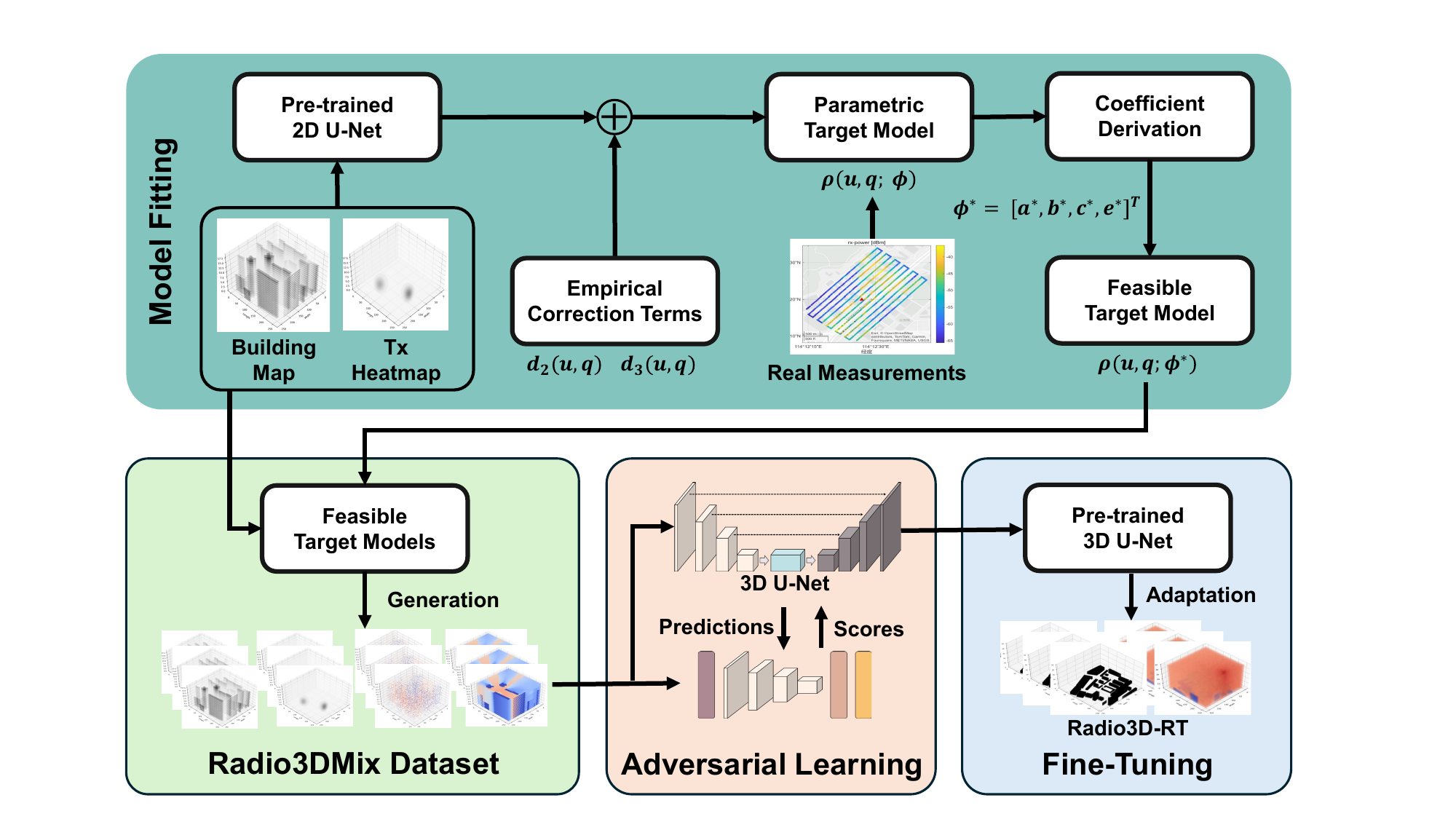}
    \caption{Diagram of the proposed RadioGen3D framework.}
    \label{fig:diagram}
\end{figure*}

In this section, we first present an efficient data synthesis approach for 3D radio map generation and a resultant dataset \emph{Radio3DMix}. Subsequently, a cGAN-based model training scheme is also provided. The diagram of the proposed RadioGen3D is illustrated in Fig. \ref{fig:diagram}.

\subsection{Efficient Synthesis of Radio3DMix Dataset}\label{sec-data-synthesis}

\subsubsection{Feature Vector Construction}
As aforementioned, the complete feature vector $\mathbf{X}_{\mathsf{full}}$ is composed of the geographical environment, the transmitter information, and sparse measurements, which are sequentially detailed as below.

\begin{itemize}
    \item \textbf{Geographical Environment:} In order to simulate the dense building environment of a real city, we select $200$ distinct regions within Beijing urban area and extract their corresponding building maps from the city map using ArcGIS software. The physical dimensions of each building map are $256 \times 256 \times 20 m^3$, i.e. $W = D = 256$ and $H = 20$. These raw building maps are then converted into voxel grids with a uniform spatial resolution of $1$ m per voxel, yielding binary occupancy tensors $\mathbf{E}$ of size $256 \times 256 \times 20$ per map. Consequently, we obtain $N_{\mathsf{E}} = 200$ building tensors, which collectively form the geographical environment part of the dataset.
    \item \textbf{Transmitter Information:} As aforementioned in Sec. \ref{communication-model}, the transmitter matrix $\mathbf{T}$ comprises the 3D locations $\mathbf{q}_i=[x_i,y_i,z_i]^T$ and transmit powers $P_i$ of all transmitters $i \in \mathcal{T}$. Specifically, given a transmitter set $\mathcal{T}$, the corresponding transmitter matrix is constructed as
\begin{align}
\mathbf{T}_{\mathbf{q}_i} = 
    \begin{cases}
    P_i, &\text{if}\ i \in \mathcal{T},\\
    0, &\text{otherwise}.
    \end{cases}
\end{align}
In this manner, we generate a total of $N_{\mathsf{T}} = 2000$ transmitter sets, each comprising two transmitters, through a randomization process and construct the corresponding transmitter matrices.


\item \textbf{Sparse Measurements:} A location set $\mathcal{S}$ is first generated by uniform sampling over the 3D space $\mathcal{R}$ with sampling rate $\xi<1$. The sparse measurement matrix is then constructed by
\begin{align}\label{measurements}
\mathbf{S}_{\mathbf{u}_j} = 
    \begin{cases}
    \mathbf{P}_{\mathbf{u}_j}, &\text{if}\ {\mathbf{u}_j} \in \mathcal{S},\\
    0, &\text{otherwise},
    \end{cases}
\end{align}
given the ground-truth label $\mathbf{P}$. Note that sparse measurement matrix is directly constructed from the label following \eqref{measurements} instead of being prepared in advance.

After introducing the feature vector, we present in the following the approach of constructing the ground-truth label $\mathbf{P}$ for each data sample.
\end{itemize}

\begin{table*}
    \centering
    \renewcommand{\arraystretch}{1.5}
    \caption{Comparison of Radio Map Datasets} 
    \label{table1}
    \begin{tabular}{|c|c|c|c|c|c|c|}
    \hline
    Properties & Radio3DMix & Radio3D-RT & UrbanRadio3D\cite{wang2025radiodiff} & SpectrumNet\cite{zhang2024generative} & RadioGAT\cite{li2024radiogat} & RadioMapSeer\cite{levie2021radiounet} \\
    \hline
    Dataset Size & 50k & 5k & 140.2k & 300k & 21k & 56k \\ 
    \hline
    Height & 20 & 50 & 20 & 3 & 1 & 1 \\
    \hline
    Map Size & 256$\times$256 & 256$\times$256 & 256$\times$256 & 128$\times$128 & 200$\times$200 & 256$\times$256 \\
    \hline
    Resolution & 1 m / voxel & 1 m / voxel & 1 m / voxel & 10 m / pixel & 5 m / pixel & 1 m / pixel\\
    \hline
    Number of Tx & 2 & 1 & 1 & 4 & 3 & 1\\
    \hline
    \end{tabular}
\end{table*}

\subsubsection{Label Construction via Data Synthesis}
Here we present an efficient data synthesis method for 3D radio map generation based on the principle of transfer learning. 
The core of our approach is to adapt a model pre-trained for 2D RME to the 3D domain using a minimal amount of 3D RSS samples, which can be acquired through real measurements or high-accuracy ray-tracing simulations. In this work, we use a few sets of real measurements collected at 2.4 GHz by a self-developed UAV-based platform to capture the characteristics of signal propagation in real-world scenarios \cite{zhonghao}. 
Based on the real measurements, and their corresponding transmitter and environment information, we construct a small trial dataset $\mathcal{D}_{\mathsf{c}}$ comprising $N_{\mathsf{c}}$ data samples, where the real measurements are used as masked labels, denoted as $\widetilde{\mathbf{P}}$. 
With such a trial dataset, the process of data synthesis is specified as follows.
\begin{itemize}
    \item \textbf{Predicted Map Generation:} Given a 3D radio map data sample $([\mathbf{T},\mathbf{S},\mathbf{E}],\widetilde{\mathbf{P}})$, a 2D feature vector corresponding to height $z$ is built as 
    \begin{align}
        \mathbf{X}_z = [\mathrm{Proj}(\mathbf{T}), \mathbf{E}_z],
    \end{align}
    where the $\mathrm{Proj}(\cdot)$ function represents a projection operation along the $Z$ direction, and $\mathbf{E}_z$ gives the 2D building matrix corresponding to height $z$. Then by applying a pre-trained model $G_{\mathsf{pre}}(\cdot;\boldsymbol{\theta}_{\mathsf{pre}})$, e.g., the U-Net model in \cite{levie2021radiounet}, we are able to calculate a predicted 2D radio map as 
    \begin{align}\label{2d-generation}
        \widehat{\mathbf{P}}_z = G_{\mathsf{pre}}(\mathbf{X}_z;\boldsymbol{\theta}_{\mathsf{pre}}).
    \end{align}
    Concatenating 2D radio maps of all heights gives the predicted 3D radio map $\widehat{\mathbf{P}} = [\widehat{\mathbf{P}}_1,\dots, \widehat{\mathbf{P}}_H]^T$.
    \item \textbf{Target Model Construction:} After obtaining $\widehat{\mathbf{P}}$, we then construct a parametric target model as
    \begin{align}
        \rho(\mathbf{u}, \mathbf{q}; {\boldsymbol{\phi}}) = &a + b\cdot\log\left(d_3(\mathbf{u}, \mathbf{q})\right) + c\cdot\log\left(d_2(\mathbf{u}, \mathbf{q}) \right) \nonumber\\
        &+e\cdot \widehat{\mathbf{P}}_{\mathbf{u}}, 
    \end{align}
    to describe the log-scale RSS at arbitrary locations $\mathbf{u}\in\mathcal{R}$,  where $\mathbf{q}$ gives the location of transmitter, the vector ${\boldsymbol{\phi}} = [a, b ,c, e]^T$ is the aggregation of the undetermined coefficients, and $\widehat{\mathbf{P}}_{\mathbf{u}}$ represents the value of matrix $\widehat{\mathbf{P}}$ at $\mathbf{u}=[x,y,z]^T$. The functions $d_3(\mathbf{u}, \mathbf{q})$ and $d_2(\mathbf{u}, \mathbf{q})$ denote the 3D spatial distance and 2D distance over the X-Y plane, respectively, as shown in Fig. \ref{fig:antenna polarization}. The principle for incorporating these empirical correction terms is given in Remark \ref{remark1}.
    \item \textbf{Model Coefficient Determination:} To obtain optimal coefficients ${\boldsymbol{\phi}^*} = [a^*, b^*, c^*, e^*]^T$ that minimize the discrepancy between the results calculated by the target model and the values in $\widetilde{\mathbf{P}}$, we formulate the coefficient determination as an error-minimization problem
    \begin{align}
        {\boldsymbol{\phi}^*} = \arg\min_{\boldsymbol{\phi}} \sum_{\mathbf{u}}\|\widetilde{\mathbf{P}}_{\mathbf{u}}-\rho(\mathbf{u}, \mathbf{q}; {\boldsymbol{\phi}})\|^2.
    \end{align}
    The optimization problem can be solved by the least-squares method. Note that the summation is taken over the locations $\mathbf{u}$ where $\widetilde{\mathbf{P}}_{\mathbf{u}}$ has definitions. This is because the label is constructed with real measurements, which are sparse over the region $\mathcal{R}$.
\end{itemize}
After obtaining the optimal ${\boldsymbol{\phi}}^*$, the target model $\rho(\cdot; {\boldsymbol{\phi}}^*)$ can be applied for radio map generation. Experimental results indicate that the mean squared error (MSE) between the generated results and the real measurements is less than $4$ dB, for every set of optimal coefficient. With $N_{\mathsf{c}}$ 3D data samples, we obtain $N_{\mathsf{c}}$ optimal coefficient vectors following the above optimization procedures. These solutions can be aggregated into a matrix
\begin{align}
    \boldsymbol{\Phi}^* = [\boldsymbol{\phi}_1^*, \dots, \boldsymbol{\phi}_{N_{\mathsf{c}}}^*]^T,
\end{align}
which characterizes the feasible variational range of the coefficients. This allows the construction of $L$ distinct target models $\{\rho(\cdot; {\boldsymbol{\phi}}_\ell^*)\}_{\ell=1}^L$ by oversampling the coefficient set $\boldsymbol{\Phi}^*$, with $L\gg N_c$. The coefficients $a$, $b$, $c$, and $e$ are varying independently within the intervals $[a_{\mathsf{lb}}, a_{\mathsf{ub}}]$, $[b_{\mathsf{lb}}, b_{\mathsf{ub}}]$, $[c_{\mathsf{lb}}, c_{\mathsf{ub}}]$, $[e_{\mathsf{lb}}, e_{\mathsf{ub}}]$, respectively.

By leveraging target models with these feasible coefficient combinations, a large number of 3D radio map data samples covering diverse signal propagation characteristics can be generated, forming a comprehensive synthetic 3D radio map dataset. Specifically, for a given target model $\rho(\cdot; {\boldsymbol{\phi}}_\ell^*)$, a synthetic 3D radio map sample $\widehat{\mathbf{P}}^*$ can be generated as
\begin{align}\label{synthesis}
\widehat{\mathbf{P}}_{\mathbf{u}}^* = 
    \begin{cases}
    P, &\text{if}\ \mathbf{u} = \mathbf{q},\\
    0, &\text{if}\ d_2(\mathbf{u}, \mathbf{q}) = 0,\\
    \rho(\mathbf{u}, \mathbf{p}; {\boldsymbol{\phi}}_\ell^*), &\text{otherwise}.
    \end{cases}
\end{align}
Note that \eqref{synthesis} describes a synthetic 3D radio map corresponding to a single transmitter with location $\mathbf{p}$. 
For scenarios involving multiple transmitters, the radio map for each transmitter can be generated sequentially, where the target model coefficients may vary across transmitters. As an example, in this work we select $L \!=\! 25$ target models with distinct coefficients and generate a total of $50000$ synthetic 3D radio maps, each containing two transmitters.
The resulting collection of synthetic 3D radio maps forms the dataset \emph{Radio3DMix}\footnote{Dataset available at \textit{https://github.com/JunshenChenJason/Radio3DMix}.}. A simple comparison between Radio3DMix and other representative existing radio map datasets are provided in Table \ref{table1}. While not the largest dataset, Radio3DMix yields a closer approximation of real-world signal propagation than ray-tracing datasets exploiting the comprehensive target model and real measurements, which thereby leads to a higher learning efficiency. Detailed parameters of the Radio3DMix dataset are listed in Table \ref{tab:dataset_params}.

\begin{table}[t]
\centering
\caption{Parameters of Radio3DMix Dataset}
\label{tab:dataset_params}
\renewcommand{\arraystretch}{1.5}  

\begin{tabular}{|p{4cm}<{\centering}|c|}  
\hline
\textbf{Parameter} & \textbf{Value} \\
\hline
Radio map size & $256\times 256\times 20\ \text{m}^3$ \\
\hline
Radio map resolution & $1\times 1\times 1\ \text{m}^3$ \\
\hline
Center carrier frequency & $2.4\ \text{GHz}$ \\
\hline
Number of building maps & 200 \\
\hline
Number of transmitter sets per building map & \multirow{2}*{10} \\  
\hline
Number of feasible target models & 25 \\
\hline
Number of radio maps & 50000 ($200\times 10\times 25$) \\
\hline
Number of transmitters per set & 2 \\
\hline
Sparse sampling rate & 1\%--10\% \\
\hline
\end{tabular}
\end{table}

\subsubsection{Properties of Radio3DMix Dataset}
\begin{figure}[t]
\centering
    \subfloat[slices at the 1 m]
    {
        \includegraphics[width=0.29\linewidth]{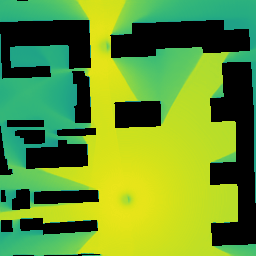}
    }\
    \subfloat[slices at the 10 m]
    {
        \includegraphics[width=0.29\linewidth]{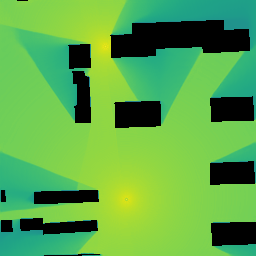}
    }\
    \subfloat[slices at the 20 m]
    {
        \includegraphics[width=0.29\linewidth]{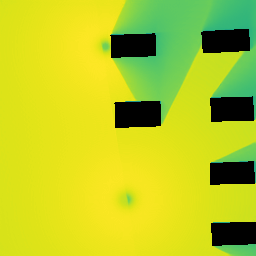}
    }
\caption{Visualized horizontal slices at different heights. The transmitter is located at height of $z=10$ m.}
\label{fig:horizontal slices at different heights}
\end{figure}

\begin{figure}[t]
\centering
    \subfloat
    {
        \includegraphics[height=1.4cm, width=1\linewidth]{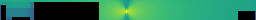}
    }\\
    \subfloat
    {
        \includegraphics[height=1.4cm, width=1\linewidth]{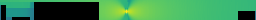}
    }\\
    \subfloat
    {
        \includegraphics[height=1.4cm, width=1\linewidth]{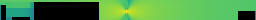}
    }
\caption{Visualized vertical slices with different model coefficients.}
\label{fig:vertical slices at different parameters}
\end{figure}

To demonstrate the capability of the target model in capturing essential 3D signal propagation characteristics, we show some representative radio maps from Radio3DMix in Fig. \ref{fig:horizontal slices at different heights} and Fig. \ref{fig:vertical slices at different parameters}.  As shown in Fig. \ref{fig:vertical slices at different parameters}, the vertical slices passing through the transmitter location $\mathbf{p}$ exhibit distinct 3D channel fading and antenna polarization characteristics under different coefficient combinations. The 3D channel fading results in weaker signal strengths at heights of $z=1$ m and $z=20$ m, compared to the signal strength at $z=10$ m where the transmitter is located. Furthermore, the horizontal slices in Fig. \ref{fig:horizontal slices at different heights} indicate that the target model also effectively captures 2D propagation characteristics. This capability is inherited from the pre-trained 2D U-Net, which serves as an approximator of 2D ray-tracing \cite{levie2021radiounet,zhang2023rme,zhang2024fast}.

\begin{figure}[t]
    \centering
    \includegraphics[width=0.85\linewidth]{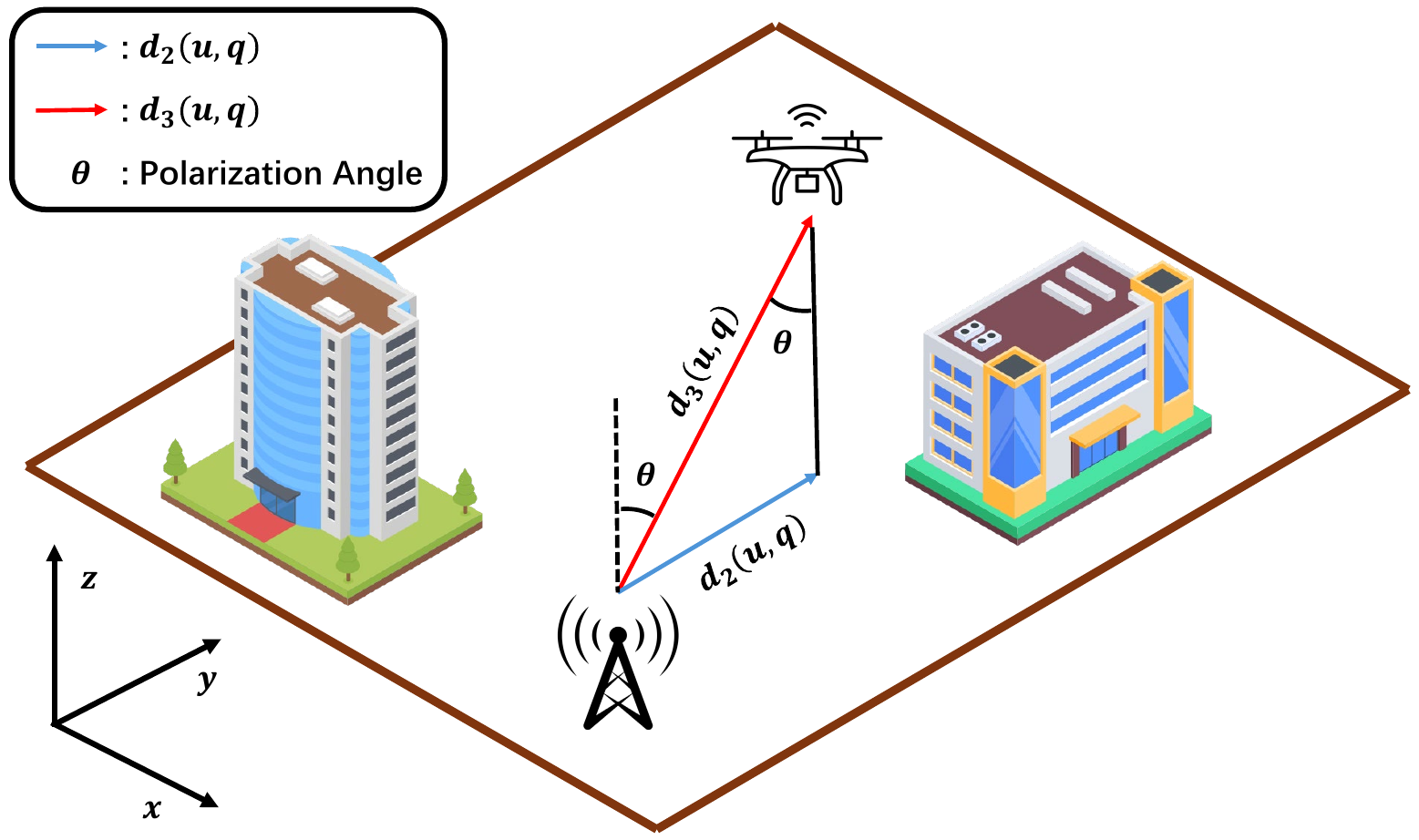}
    \caption{Illustration of 3D channel fading under vertical polarization.}
    \label{fig:antenna polarization}
\end{figure}

\begin{rem}[Effects of 3D Channel Fading and Antenna Polarization]\label{remark1}
    The major limitation of using a pre-trained 2D model for radio map estimation is its inability to capture 3D signal propagation characteristics, particularly 3D channel fading and antenna polarization effects. First, as illustrated in Fig. \ref{fig:antenna polarization}, the pathloss in log-scale at the receiver is proportional to $\log\left(d_3(\tau, \kappa)\right)$, which aligns exactly with the term $b\cdot\log\left(d_3(\mathbf{u}, \mathbf{q})\right)$ in the target model. Second, according to \cite{balanis2016antenna}, under vertical polarization conditions, the antenna gain in log-scale at polarization angle $\theta$ can be expressed as
\begin{align}
    \rho_{\mathsf{ap}} =& \log\left(\cos(\frac{\pi}{2}\cos\theta)^2\right) \nonumber\\
    \approx& \log\left(\sin^5(\theta )\right)\nonumber\\
    =&5 \left[\log\left(d_2(\mathbf{u}, \mathbf{q})\right) - \log\left(d_3(\mathbf{u}, \mathbf{q})\right)\right],
\end{align}
which can be approximated by a linear combination of $\log\left(d_2(\mathbf{u}, \mathbf{q})\right)$ and $\log\left(d_3(\mathbf{u}, \mathbf{q})\right)$. Therefore, by incorporating the empirical correction terms $\log\left(d_2(\mathbf{u}, \mathbf{q})\right)$ and $\log\left(d_3(\mathbf{u}, \mathbf{q})\right)$, the target model becomes capable of representing 3D channel fading and antenna polarization effects after optimization \cite{sun2025support}. 
\end{rem}

\subsection{cGAN-based 3D Radio Map Estimation}
Following the construction of the synthetic Radio3DMix dataset, we proceed to introduce a 3D model training scheme by exploiting the cGAN framework. The model architecture, data pre-processing, and training procedures are presented in the following.

\subsubsection{Model Selection}
To facilitate 3D radio map estimation, we adopt a 3D U-Net as the generator in this work. This is motivated by the following reasons.  First, the 3D U-Net has been widely and successfully applied in 3D image-to-image generation tasks and exhibits favorable performance in RME \cite{levie2021radiounet,wang2024radiodiff,wang2025radiodiff}, demonstrating its strong capability in this domain. Second, the pre-trained model $G_{\mathsf{pre}}(\cdot;\boldsymbol{\theta}_{\mathsf{pre}})$ used in \eqref{2d-generation} is a 2D U-Net. The inherent structural similarity between the 2D and 3D U-Nets is expected to enhance both training efficiency and final performance. 

\subsubsection{Data Pre-processing}
To ensure compatibility with the U-Net generator, the data are processed through the following transformations.

\textbf{Image Transformation:} As the 3D RME is formulated as an image-to-image generation task, both input features and the label need to be converted into images. Through normalization and quantization across the whole dataset, each data sample $(\mathbf{X}, \mathbf{P})$ is transformed into grayscale image pairs, denoted as $(\mathbf{X}^{\mathsf{img}}, \mathbf{P}^{\mathsf{img}})$. Specifically, the input features are represented as a 4D tensor $\mathbf{X}^{\mathsf{img}}\in \mathbb{R}^{C\times W\times D\times H}$, with $C$ being the number of image channels.

\textbf{Transmitter Heatmap Encoding:} Through image-form transformation, the transmitter information $\mathbf{T}$ is first converted into a sparse, multi-hot tensor by mapping each transmitter to a single voxel at its corresponding 3D coordinate. Although employed in existing works \cite{levie2021radiounet,wang2025radiodiff}, it is well recognized that such a representation poses two main difficulties due to its extreme sparsity. First, the multi-hot transmitter information are susceptible to the vanishing problem after successive convolution and downsampling operations, leading to unstable feature extraction and thus impeding effective learning. Second, the significant sparsity discrepancy between the transmitter tensor and other dense input features, e.g., measurements and high-resolution environment, makes it difficult for the model to learn the correlation between the transmitter locations and the resultant signal propagation patterns in the label. To address this limitation, in this work we propose to encode each transmitter into a 3D Gaussian heatmap centered at its location with a ternary Gaussian distribution, given as
    \begin{align}
    \widetilde{\mathbf{T}}_{i,\mathbf{u}}^{\mathsf{img}} = P_i\cdot\exp\left(\!-\frac{\left(\mathbf{u}-\mathbf{q}_i \right)^T \mathbf{\Sigma}_i^{-1} \left(\mathbf{u}-\mathbf{q}_i \right)}{2} \!\right),
    \end{align}
    where $\mathbf{\Sigma}_i^{-1}$ is the covariance matrix, denoted as a diagonal matrix
    \begin{align}\label{covariance-matrix}
    \mathbf{\Sigma}_i^{-1} = 
    \begin{bmatrix} 
        \frac{1}{\sigma_z^2} & 0 & 0 \\ 0 & \frac{1}{\sigma_{xy}^2} & 0 \\ 0 & 0 & \frac{1}{\sigma_{xy}^2}
    \end{bmatrix},
    \end{align}
    with $\sigma_z = D \times 0.1$ and $\sigma_{xy} = \min(H, W) \times 0.05$. Subsequently, inter-channel normalization is applied as
    \begin{align}
    \mathbf{T}_{i,\mathbf{u}}^{\mathsf{img}} = \frac{\widetilde{\mathbf{T}}_{i,\mathbf{u}}^{\mathsf{img}}}{\max \{ \widetilde{\mathbf{T}}_{i,\mathbf{u}}^{\mathsf{img}}, \forall i, \mathbf{u} \}}.
    \end{align}
    Finally, the Gaussian heatmaps corresponding to all transmitters are aggregated as a tensor
    \begin{align}
    \mathbf{T}^{\mathsf{img}} = [\mathbf{T}^{\mathsf{img}}_{1}, \mathbf{T}^{\mathsf{img}}_{2}, \dots, \mathbf{T}^{\mathsf{img}}_T]^T,
    \end{align}
    which gives the input feature of transmitter information.

\subsubsection{cGAN-based Model Training}
\begin{figure*}[t]
    \centering
    \includegraphics[width=0.66\linewidth]
    {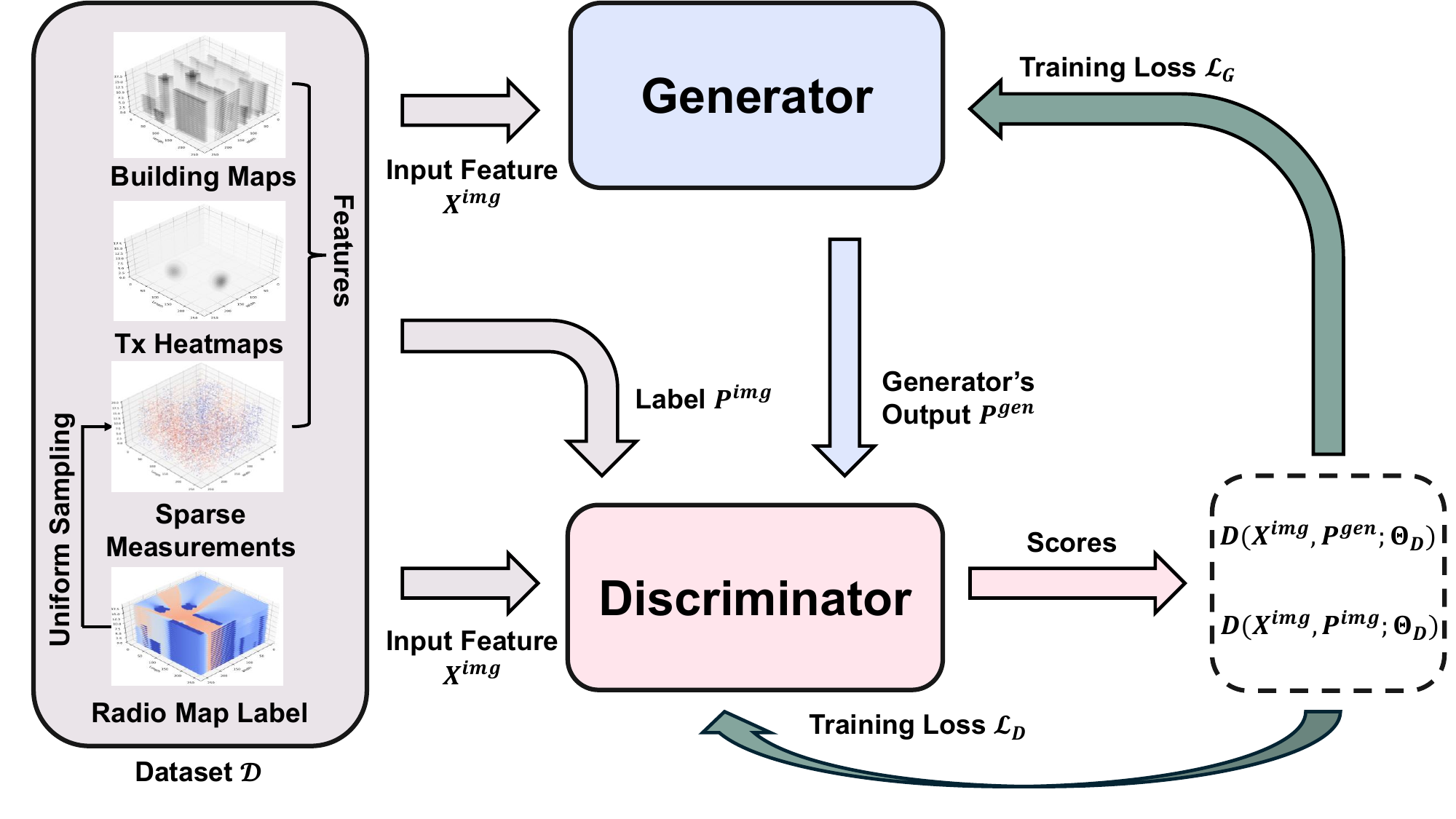}
    \caption{The proposed cGAN-based training process in RadioGen3D}
    \label{fig:Structure of cGAN}
\end{figure*}

As aforementioned, a 3D U-Net model is adopted as the generator. To achieve effective model training, we employ conditional GAN, where a discriminator $D(\cdot;\boldsymbol{\Theta}_D)$ is introduced to enhance the learning performance of the generator via adversarial training. The training workflow of cGAN is depicted in Fig. \ref{fig:Structure of cGAN}. Different from standard GANs, where the discriminator receives only the generated image and the ground-truth label, the discriminator in cGAN distinguishes between two joint distributions: 1) the joint distribution of the input features and the label $P(\mathbf{X}^{\mathsf{img}}, \mathbf{P}^{\mathsf{img}})$, and 2) the joint distribution of the input features and the generated output $P(\mathbf{X}^{\mathsf{img}}, {\mathbf{P}}^{\mathsf{gen}})$. The specific model structures and implementation details of both the generator and discriminator are described in the following.
 
\begin{figure*}[htbp]
    \centering
    \subfloat
    {
        \includegraphics[width=0.45\linewidth]
        {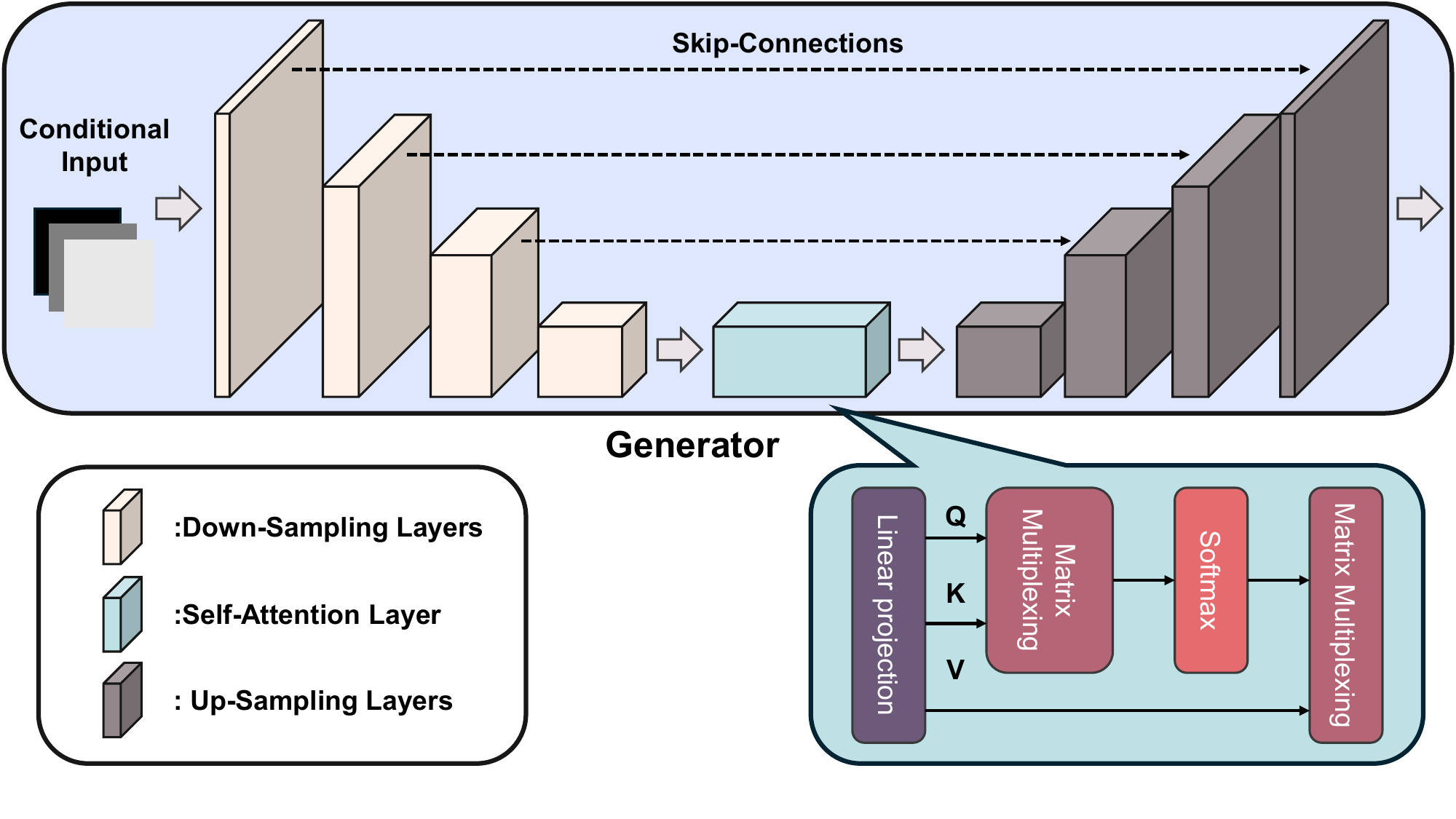}
    }
    {
        \includegraphics[width=0.45\linewidth]
        {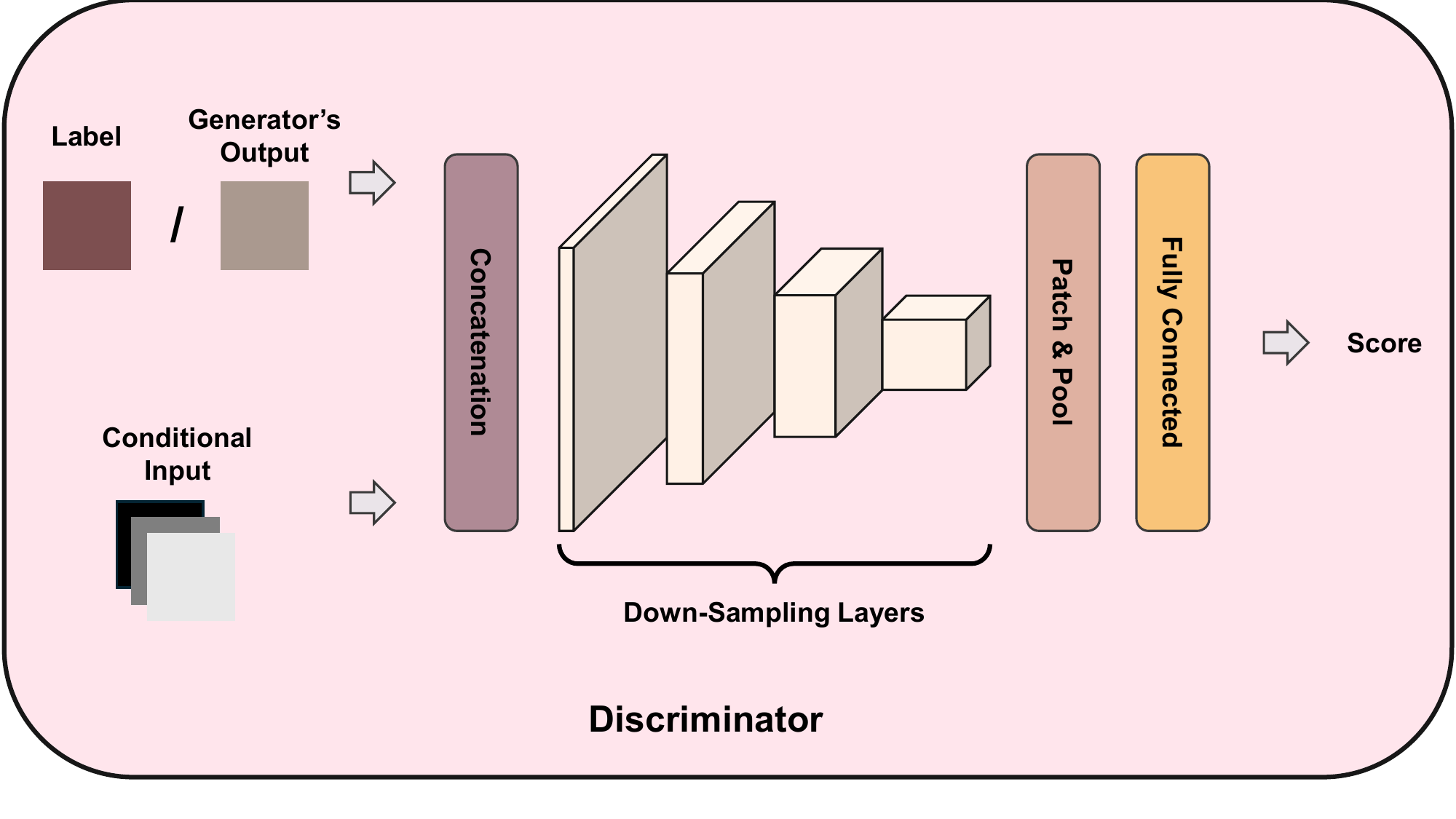}
    }
    \caption{Model Structure of Generator and Discriminator}\vspace{-0.3cm}
    \label{fig:Structure of Generator and Discriminator}
\end{figure*}

\textbf{Generator:} A 3D U-Net is adopted as the generator to map the input features $\mathbf{X}^{\mathsf{img}}$ to the 3D radio map label $\mathbf{P}^{\mathsf{img}}$. Within a 3D U-Net, the encoder progressively reduces spatial resolution through downsampling to capture multi-scale semantic features, while the decoder performs upsampling and incorporates high-resolution local information from the encoder via skip connections, thereby maintaining structural and semantic fidelity in the output.
The specific structures of both the generator and discriminator are illustrated in Fig. \ref{fig:Structure of Generator and Discriminator}. In the U‑Net generator, each downsampling module within the encoder uses a 3D convolution layer $\mathrm{Conv3D}(\cdot)$ with a kernel designed to avoid reducing the height dimension $D$, thereby preserving higher-resolution information along the vertical direction. This design is crucial for capturing altitude-dependent signal variations and allows the generator to adapt automatically to data samples of different heights. Each downsampling module, which integrates a normalization layer $\mathrm{GN}(\cdot)$ and an activation layer $\mathrm{LeakyReLU}(\cdot)$, can be expressed as
\begin{align}
    \mathbf{X}^{\mathsf{out}} = \mathrm{LeakyReLU}\left(\mathrm{GN}\left(\mathrm{Conv3D}(\mathbf{X}^{\mathrm{in}})\right)\right).
\end{align}
In the decoder, image resolution is restored using normalization and activation layers similar to those in the encoder, together with a transposed convolution. Following upsampling, feature maps are concatenated with high‑resolution local information from the corresponding encoder layer via skip connections. A subsequent convolution is then applied to fuse the concatenated features.

Moreover, in order to capture long-range 3D dependencies, a self-attention module is incorporated into the bottleneck of the U-Net. Let $\mathbf{X}^{\mathsf{enc}}$ denote the output feature map of the final encoder layer, linear projections are first conducted as
\begin{align}
    [\mathbf{Q}, \mathbf{K}, \mathbf{V}] = [\mathbf{W}_q \mathbf{X}^{\mathsf{enc}},  \mathbf{W}_k \mathbf{X}^{\mathsf{enc}},  \mathbf{W}_v \mathbf{X}^{\mathsf{enc}}],
\end{align}
where $\mathbf{W}_q$, $\mathbf{W}_k$ and $\mathbf{W}_v$ are trainable projection tensors. The self-attention output is computed as
\begin{align}
    \mathbf{X}^{\mathsf{atten}} = \mathrm{softmax}\left(\mathbf{Q}\mathbf{K}^T \right)\mathbf{V}.
\end{align}
Finally, the attention output is fused with $\mathbf{X}^{\mathsf{enc}}$ via a residual connection, given as
\begin{align}
    \mathbf{X}^{\mathsf{atten}} \leftarrow \gamma \cdot \mathbf{X}^{\mathsf{atten}} + \mathbf{X}^{\mathsf{enc}},
\end{align}
where $\gamma$ is a learnable scaling parameter.

\textbf{Discriminator:} The discriminator is designed to distinguish between the ground-truth labels and the output predictions from the generator, given the same conditional inputs. This adversarial feedback guides the generator toward producing more realistic outputs. During training, the discriminator evaluates two types of input combinations for each data sample: 
\begin{itemize}
    \item $[\mathbf{X}^{\mathsf{img}}, \mathbf{P}^{\mathsf{img}}]$: Concatenation of conditional information $\mathbf{X}^{\mathsf{img}}$ and the label $\mathbf{P}^{\mathsf{img}}$;
    \item $[\mathbf{X}^{\mathsf{img}}, \mathbf{P}^{\mathsf{gen}}]$: Concatenation of conditional information $\mathbf{X}^{\mathsf{img}}$ and the prediction ${\mathbf{P}}^{\mathsf{gen}}$ from the generator. 
\end{itemize}
For the model structure, the discriminator consists of multi-layer 3D convolutions to extract spatial features. The final layer performs patch-level discrimination to enable the assessment of local signal consistency and physical plausibility across the 3D volume, rather than relying solely on the global statistics.

\textbf{Loss Functions:} To achieve effective training and enhance the capability of both the generator and discriminator, the loss functions of are designed as follows. Firstly, for the generator, its loss function $\mathcal{L}_{G}$, i.e., the function $\mathrm{Dist}(\cdot)$ in \eqref{problem-formulation}, is defined as the weighted sum of an adversarial loss $\mathcal{L}_{adv}$ and a voxel-wise loss $\mathcal{L}_{L1}$, given as
\begin{align}\label{generator-loss}
    \mathcal{L}_{G} = \alpha_{\mathsf{adv}}\mathcal{L}_{\mathsf{adv}} + \lambda_{\mathsf{L1}}\mathcal{L}_{\mathsf{L1}},
\end{align}
where 
\begin{align}
    \mathcal{L}_{\mathsf{adv}} = \frac{1}{2}\mathbb{E}\left[\left(D\left(\mathbf{X}^{\mathsf{img}}, {\mathbf{P}}^{\mathsf{gen}}; \boldsymbol{\Theta}_D \right)-1 \right)^2\right],
\end{align}
and 
\begin{align}
    \mathcal{L}_{\mathsf{L1}} = \mathbb{E}\left[\|\mathbf{P}^{\mathsf{img}}- {\mathbf{P}}^{\mathsf{gen}} \|\right].
\end{align}
The weights $\alpha_{\mathsf{adv}}$ and $\lambda_{\mathsf{L1}}$ can be dynamically adjusted during training to enhance training stability. For the voxel-wise loss, we adopt the $L1$ loss instead of the $L2$ loss as it is proven to be less sensitive to outliers and is able to better preserve boundary features of local signals in radio maps.

For the discriminator, the loss function is composed of two adversarial losses and a regularization loss, given as
\begin{align}
    \mathcal{L}_{D} = &\frac{1}{2}\mathbb{E}\left[\left(D\left(\mathbf{X}^{\mathsf{img}}, \mathbf{P}^{\mathsf{img}}; \boldsymbol{\Theta}_D \right)-1 \right)^2\right] +  \nonumber\\
    &\frac{1}{2}\mathbb{E}\left[\left(D\left(\mathbf{X}^{\mathsf{img}}, {\mathbf{P}}^{\mathsf{img}}; \boldsymbol{\Theta}_D \right) \right)^2\right] + L_{\mathsf{R}},
\end{align}
where
\begin{align}
    L_{\mathsf{R}} = \frac{\gamma}{2}\mathbb{E}\left[\left|\nabla_{\mathbf{P}^{\mathsf{img}}} D\left(\mathbf{X}^{\mathsf{img}}, \mathbf{P}^{\mathsf{img}}; \boldsymbol{\Theta}_D \right)\right|^2\right],
\end{align}
denotes the regularization loss, with $\gamma$ being the regularization coefficient.

\section{Experimental Results}\label{sec:experiments}
In this section, we evaluate the effectiveness of the proposed RadioGen3D through experiments. In addition to evaluating the model on  Radio3DMix, we also adapt it for a ray-tracing dataset via fine-tuning to facilitate a fair comparison with existing baselines. The experimental configurations, evaluation metrics and results are detailed in the sequel.

\subsection{Experimental Configurations}
To give a comprehensive evaluation on RadioGen3D, we establish three experimental configurations in the following, each corresponding to one of the three input feature vectors described in Sec. \ref{sec-problem-formulation} and modeled in \eqref{model1}, \eqref{model2}, and \eqref{model3}.
\begin{itemize}
    \item \textbf{Sparse-and-Transmitter}: The transmitter information, sparse measurements, and environment information are all incorporated into the feature vector $\mathbf{X} = [\mathbf{T}, \mathbf{S}, \mathbf{E}]$.
    \item \textbf{Sparse-Only}: Only the sparse measurements and the environment information are contained in the feature vector, given as $\mathbf{X} = [\mathbf{S}, \mathbf{E}]$. 
    \item \textbf{Transmitter-Only}: Only the transmitter and environment information are contained in the feature vector, given as $\mathbf{X} = [\mathbf{T}, \mathbf{E}]$. 
\end{itemize}
In all the three configurations above, the feature vector $\mathbf{X}$ is further transformed into the image form $\mathbf{X}^{\mathsf{img}}$ as the input to the generator and discriminator.

The above configurations cover practical scenarios where the transmitter information or sparse measurements can be inaccurate or unavailable, as discussed in Sec. \ref{sec-problem-formulation}. Thus it is essential for the model to handle different feature combinations effectively for robust RME. In addition, the entire dataset is partitioned into training (60\%), validation (20\%), and testing (20\%) sets.

The default settings of model training are described as follows unless specified otherwise. All models were trained on an NVIDIA RTX $5090$ GPU with a batch size of $8$ for $100$ epochs. The learning rates are set to $2\times10^{-4}$ and $1\times10^{-4}$ for the generator and discriminator, respectively. The weight $\lambda_{\mathsf{L1}}$ for the $L1$ loss in \eqref{generator-loss} is fixed at $100$. A phased scheduling strategy is employed for the adversarial loss weight $\alpha_{\mathsf{adv}}$. It is set to $0$ for the first $15$ epochs, increased linearly over the subsequent $10$ epochs until reaching $0.1$, and maintained thereafter. This prevents the discriminator from overwhelming the generator prematurely.

\subsection{Performance Metrics}
To quantify the performance of the proposed RadioGen3D, we apply a few metrics that are widely adopted in RME and image reconstruction tasks, including the root mean square error (RMSE), normalized mean square error (NMSE), structural similarity index measure (SSIM), and peak signal-to-noise ratio (PSNR). The detailed definitions and interpretations of these metrics are described below.

\subsubsection{RMSE}
the RMSE is the square root of the mean squared error (MSE), which is defined as 
\begin{align}
    E_{\mathsf{RMSE}} = \sqrt{\frac{1}{|\mathcal{V}|}\sum_{v\in \mathcal{V}}\left({\mathbf{P}}^{\mathsf{gen}}_v-\mathbf{P}^{\mathsf{img}}_v\right)^2},
\end{align}
where $v\in \mathcal{V}$ indexes a voxel within the volumetric space $\mathcal{V}$ corresponding to the region $\mathcal{S}$.

\subsubsection{NMSE}
NMSE is a scale-invariant metric derived from MSE that normalizes MSE by the energy of the ground-truth map. In RME, it is defined as the ratio of the squared reconstruction error to the squared magnitude of the ground-truth label, given by
\begin{align}
    E_{\mathsf{NMSE}} = \frac{\sum_{v\in \mathcal{V}}\left({\mathbf{P}}^{\mathsf{gen}}_v-\mathbf{P}^{\mathsf{img}}_v\right)^2}{\sum_{v\in \mathcal{V}}\left(\mathbf{P}^{\mathsf{img}}_v\right)^2},
\end{align}

\begin{figure*}[t]
\centering
    {
        \includegraphics[width=0.15\linewidth]{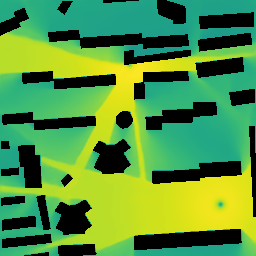}
    }\
    {
        \includegraphics[width=0.15\linewidth]{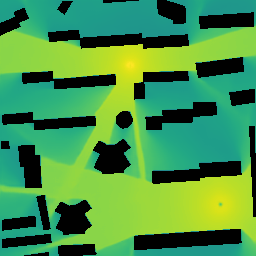}
    }\
    {
        \includegraphics[width=0.15\linewidth]{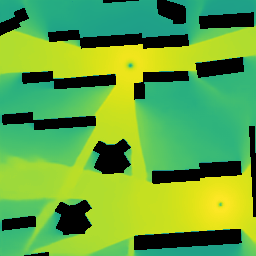}
    }\
    {
        \includegraphics[width=0.15\linewidth]{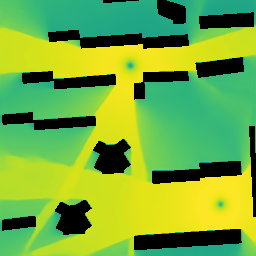}
    }\
    {
        \includegraphics[width=0.15\linewidth]{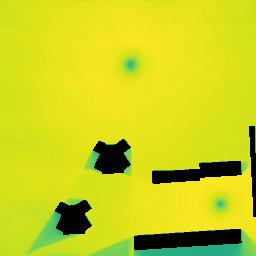}
    }\\
    \subfloat[slices at the 1 m]
    {
        \includegraphics[width=0.15\linewidth]{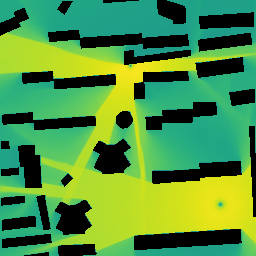}
    }\
    \subfloat[slices at the 5 m]
    {
        \includegraphics[width=0.15\linewidth]{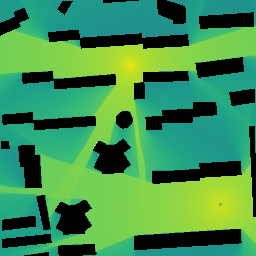}
    }\
    \subfloat[slices at the 10 m]
    {
        \includegraphics[width=0.15\linewidth]{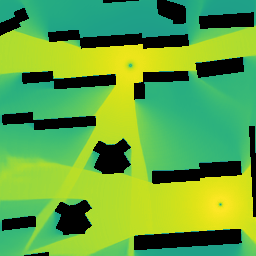}
    }\
    \subfloat[slices at the 15 m]
    {
        \includegraphics[width=0.15\linewidth]{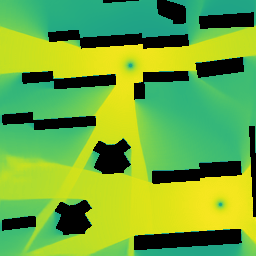}
    }\
    \subfloat[slices at the 20 m]
    {
        \includegraphics[width=0.15\linewidth]{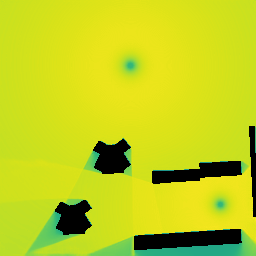}
    }
\caption{Comparison between predicted results (top row) and ground truth (bottom row) at different heights}
\label{fig:Comparison at different heights}
\end{figure*}

\subsubsection{SSIM}
SSIM is a perceptual metric that quantifies image similarity by comparing local patterns of luminance, contrast, and structure \cite{wang2004image}. Due to its capability of capturing the fidelity of local spatial structures and fading patterns, SSIM is considered particularly suitable for evaluating the quality of RME \cite{wang2025beamckm}. In the implementation, SSIM is computed locally over sliding voxel windows and then averaged over the entire 3D volume to obtain an overall score. Let ${\mathbf{P}}^{\mathsf{gen}}(w)$ and $\mathbf{P}^{\mathsf{img}}(w)$ denote the predicted and ground-truth radio maps within a local window $w$, respectively. The local SSIM is defined as
\begin{align}
        &E_{\mathsf{SSIM}}\left({\mathbf{P}}^{\mathsf{gen}}(w), \mathbf{P}^{\mathsf{img}}(w)\right) \nonumber\\
        =& \frac{\left(2\mu_1(w)\mu_2(w)+C_1\right) \left(2\Sigma(w)+C_2\right)}{\left(\mu_1^2(w)+\mu_2^2(w)+C_1\right) \left(\sigma_1^2(w)+\sigma_2^2(w)+C_2\right)},
\end{align}
where $\mu_1(w)$,  $\sigma_1^2(w)$ are the mean and variance of ${\mathbf{P}}^{\mathsf{gen}}(w)$, $\mu_2(w)$,  $\sigma_2^2(w)$ are the mean and variance of ${\mathbf{P}}^{\mathsf{img}}(w)$, and $\Sigma(w)$ denotes the covariance with respect to ${\mathbf{P}}^{\mathsf{gen}}(w)$ and ${\mathbf{P}}^{\mathsf{img}}(w)$. Additionally, The constants $C_1$ and $C_2$ are stabilization terms to avoid division by zero, defined as
\begin{align}
    C_1 = \left(K_1R\right)^2,\ C_2 = \left(K_2R\right)^2,
\end{align}
where $R$ denotes the dynamic range of the signal values. For signals normalized to $[0, 1]$, $R = 1.0$, $K_1 = 0.01$ and $K_2 = 0.03$ are empirically determined constants following the standard SSIM formulation \cite{wang2004image}.

\subsubsection{PSNR}
PSNR is a widely used quantitative metric for evaluating the fidelity of reconstructed signals by measuring the ratio between the maximum possible signal power and the power of the distortion introduced during reconstruction \cite{hore2010image}. It is typically expressed in dB and emphasizes the relative magnitude of reconstruction image with respect to the signal dynamic range. In general, a higher PSNR value corresponds to lower reconstruction error and better overall quality. 

Given that distortions in predicted radio maps are often concentrated near building boundaries, PSNR is particularly sensitive to inaccuracies in these regions \cite{wang2024radiodiff}. Consequently, PSNR is employed not only to assess the global reconstruction quality of generated radio maps, but also to indirectly evaluate the preservation of boundary details and high contrast propagation characteristics. PSNR is defined based on MSE as
\begin{align}
    E_{\mathsf{PSNR}} = 10\log_{10}\left(\frac{R^2}{E_{\mathsf{MSE}}}\right),
\end{align}
where
\begin{align}
    E_{\mathsf{MSE}} = \frac{1}{|\mathcal{V}|}\sum_{v\in \mathcal{V}}\left({\mathbf{P}}^{\mathsf{gen}}_v-\mathbf{P}^{\mathsf{img}}_v\right)^2.
\end{align}

\subsection{Performance Evaluation}
Here we provide the experimental results, including the performance evaluation of the proposed RadioGen3D and its comparisons with existing baselines \cite{wang2025radiodiff,liu2025genradio}.

\begin{table*}[t]
\centering
\caption{Performance comparison of different methods under various sampling rate}
\label{tab2}
    \begin{tabular}{c c c c c c c c c}
        \toprule
        \midrule
        Sparse sampling rate & \multicolumn{8}{c}{1\%} \\
        \cmidrule{2-9}
        Configuration & \multicolumn{4}{c}{Sparse-and-Transmitter} & \multicolumn{4}{c}{Sparse-Only} \\
        \cmidrule(lr){2-5}\cmidrule(lr){6-9} 
        Metrics & RMSE & NMSE & SSIM & PSNR & RMSE & NMSE & SSIM & PSNR\\
        \midrule
        RadioGen3D & 0.0056 & 0.0032 & 0.9965 & 45.00 & 0.0062 & 0.0041 & 0.9861 & 44.58 \\
        RadioGen3D (Fine-tuning) & 0.0278 & 0.0199 & 0.9407 & 31.09 & 0.0341 & 0.0265 & 0.9334 & 29.33 \\
        RadioGen3D (Train on RT Data) & 0.0317 & 0.0245 & 0.9371 & 29.76 & 0.0324 & 0.0257 & 0.9316 & 28.97 \\
        GenRadio \cite{liu2025genradio} & 0.1389 & / & / & 17.152 & / & / & / & / \\
        RME-GAN (2D) \cite{zhang2023rme} & 0.0151 & 0.0043 & / & / & / & / & / & / \\
        RadioUNet (2D, 0.5\%) \cite{levie2021radiounet} & 0.0209 & 0.0086 & / & / & / & / & / & / \\
        \midrule

        \midrule
        Sparse sampling rate & \multicolumn{8}{c}{10\%} \\
        \cmidrule{2-9}
        Configuration & \multicolumn{4}{c}{Sparse-and-Transmitter} & \multicolumn{4}{c}{Sparse-Only} \\
        \cmidrule(lr){2-5}\cmidrule(lr){6-9}
        Metrics & RMSE & NMSE & SSIM & PSNR & RMSE & NMSE & SSIM & PSNR \\
        \midrule
        RadioGen3D & 0.0046 & 0.0021 
        & 0.9973 & 46.71 & 0.0050 & 0.0027 & 0.9914 & 45.77 \\
        RadioGen3D (Fine-tuning) & 0.0250 & 0.0169 & 0.9417 & 31.37 & 0.0264 & 0.0181 & 0.9411 & 31.18 \\
        RadioGen3D (Train on RT Data) & 0.0271 & 0.0195 & 0.9409 & 31.13 & 0.0281 & 0.0204 & 0.9387 & 30.57 \\
        RadioDiff-3D \cite{wang2025radiodiff} & 0.0481 & 0.0550 & 0.8187 & 29.23 & / & / & / & / \\
        \midrule

        \midrule
        Configuration & \multicolumn{8}{c}{Transmitter-Only} \\
        \cmidrule{2-9}
        Metrics & \multicolumn{2}{c}{RMSE} & \multicolumn{2}{c}{NMSE} & \multicolumn{2}{c}{SSIM} & \multicolumn{2}{c}{PSNR}\\
        \midrule
        RadioGen3D & \multicolumn{2}{c}{0.0746} & \multicolumn{2}{c}{0.0141} & \multicolumn{2}{c}{0.9682} & \multicolumn{2}{c}{22.54} \\
        RadioGen3D (Fine-tuning) & \multicolumn{2}{c}{0.1140} & \multicolumn{2}{c}{0.3015} & \multicolumn{2}{c}{0.9080} & \multicolumn{2}{c}{18.86} \\
        RadioGen3D (Train on RT Data) & \multicolumn{2}{c}{0.1227} & \multicolumn{2}{c}{0.3138} & \multicolumn{2}{c}{0.8921} & \multicolumn{2}{c}{18.19} \\
        RadioDiff-3D & \multicolumn{2}{c}{0.0653} & \multicolumn{2}{c}{0.0534} & \multicolumn{2}{c}{0.8309} & \multicolumn{2}{c}{24.00} \\
        \midrule
        \bottomrule
        
    \end{tabular}
\end{table*}

\begin{figure}[t]
\centering
    \subfloat
    {
        \includegraphics[height = 1.4cm, width=1\linewidth]{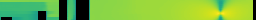}
    }\\
    \subfloat
    {
        \includegraphics[height = 1.4cm, width=1\linewidth]{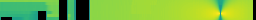}
    }\\
\caption{Comparison between the predicted result (top) and ground truth (bottom) at transmitter locations}\vspace{-0.3cm}
\label{fig:Comparison at different tx locations}
\end{figure}

\begin{table}[t]
\centering
\caption{Inference time comparison of different methods}
\label{tab3}
    \begin{tabular}{c c c}
        \toprule
        Method & Number of Height & Inference Time (sec)\\
        \midrule
        RadioGen3D & 20 & 0.06 \\
        RadioDiff-3D & 20 & 1.43 \\
        RadioDiff (2D) & 1 & 0.35 \\
        RadioUNet (2D) & 1 & 0.033 \\
        RME-GAN (2D) & 1 & 0.025 \\
        \bottomrule
    \end{tabular}\vspace{-0.2cm}
\end{table}

\subsubsection{Estimation Accuracy on Radio3DMix}
After training the 3D U-Net generator on Radio3DMix, we evaluate the estimation accuracy of the 3D U-Net using the metrics introduced in the previous subsection, which are summarized in Table \ref{tab2}. Some visualized results are also shown in Fig. \ref{fig:Comparison at different heights} and Fig. \ref{fig:Comparison at different tx locations}.

Firstly, in our experiments, the model is trained and tested under three configurations and two sparse sampling rates, to simulate various practical scenarios, as shown in Table. \ref{tab2}. Under the two configurations with sparse measurements, the proposed RadioGen3D, trained and tested on the synthetic Radio3DMix dataset, demonstrates significantly higher estimation accuracy than existing baselines that are trained and tested on ray-tracing simulated datasets. For instance, at a $1\%$ sparse sampling rate, both RMSE and NMSE values of RadioGen3D remains below 0.01, substantially outperforming other existing solutions in estimation accuracy. The potential reasons between such a significant performance boost are elaborated as follows.
\begin{itemize}
    \item \textbf{Data Sufficiency}: As presented in Sec. \ref{sec-data-synthesis}, we have generated $50000$ radio maps corresponding to $25$ distinct feasible target models, whose coefficients are derived based on real measurement by unmanned aerial vehicles (UAVs). In fact, this data synthesis framework allows for the generation of an arbitrarily large number of samples, covering all possible types of signal propagation characteristics. With such a sufficient amount of data, the generator gains strong capability, resulting in superior estimation accuracy.
    \item \textbf{Alignment between Model and Data:} More importantly, as described in Sec. \ref{sec-data-synthesis}, the Radio3DMix dataset are generated using a target model that integrates a 2D U-Net with empirical correction terms. As such, the generated data inherently embodies structural features highly compatible with U-Net architectures. This compatibility is leveraged by our 3D U-Net generator, thereby significantly improving the training efficiency and robustness.
\end{itemize}

While the data in Radio3DMix is generated through a more structured process than traditional ray-tracing, this does not imply that it is less realistic or effective. This can be validated from two aspects. Firstly, the 2D U-Net model is demonstrated as a high-fidelity approximator of ray-tracing in prior works \cite{levie2021radiounet}, thereby capturing realistic signal propagation characteristics as shown in Fig. \ref{fig:horizontal slices at different heights}. Secondly, by integrating empirical models, the proposed data synthetic approach also incorporates 3D fading and antenna polarization effects. Those features are usually missing in existing ray-tracing simulated datasets, which typically assume omnidirectional antennas. 
By exploiting such a comprehensive target model and real measurements, Radio3DMix yields a closer approximation of real-world signal propagation than ray-tracing datasets. Although the above comparison with existing baselines is inadequate due to dataset differences, such experimental results strongly demonstrates the effectiveness of our proposed method. 

Moreover, in the transmitter-only configuration, the proposed RadioGen3D does not exhibit superior performance.
This is because a single 3D U-Net model is not able to adjust to different signal propagation characteristics by only providing the transmitter and environment information. In principle, the proposed RadioGen3D relies on the sparse measurement input to infer the antenna polarization characteristics. Therefore, trained over Radio3DMix, the proposed RadioGen3D is not expected to achieve a favorable performance under the transmitter-only configuration.
Nevertheless, under such a circumstances, RadioGen3D still maintains performance comparable to the RadioDiff-3D approach, which also employs a 3D U-Net while relying on a diffusion process. In summary, the experimental results demonstrate that RadioGen3D significantly outperforms existing baselines, particularly in scenarios where sparse measurements are available.

\begin{figure*}[t]
\centering
    {
        \includegraphics[width=0.15\linewidth]{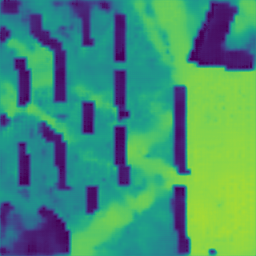}
    }\
    {
        \includegraphics[width=0.15\linewidth]{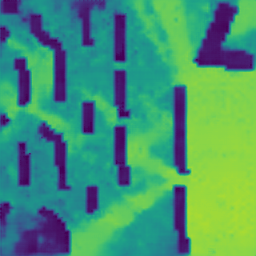}
    }\
    {
        \includegraphics[width=0.15\linewidth]{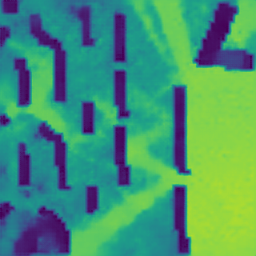}
    }\
    {
        \includegraphics[width=0.15\linewidth]{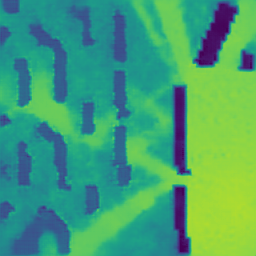}
    }\
    {
        \includegraphics[width=0.15\linewidth]{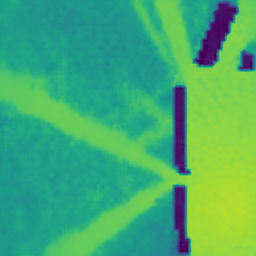}
    }\\
    \subfloat[slices at the 1 m]
    {
        \includegraphics[width=0.15\linewidth]{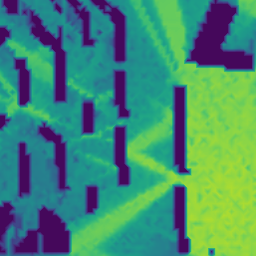}
    }\
    \subfloat[slices at the 5 m]
    {
        \includegraphics[width=0.15\linewidth]{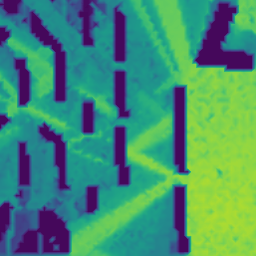}
    }\
    \subfloat[slices at the 10 m]
    {
        \includegraphics[width=0.15\linewidth]{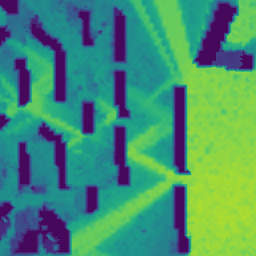}
    }\
    \subfloat[slices at the 15 m]
    {
        \includegraphics[width=0.15\linewidth]{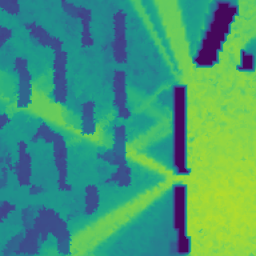}
    }\
    \subfloat[slices at the 20 m]
    {
        \includegraphics[width=0.15\linewidth]{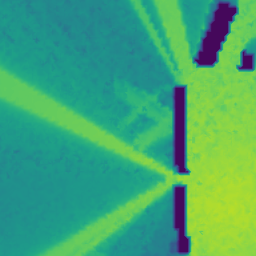}
    }
\caption{Comparison between predicted results (top row) and ground truth (bottom row) at different heights after fine-tuning.}\vspace{-0.3cm}
\label{fig:Comparison of fine-tuning experiments at different heights}
\end{figure*}

In Table \ref{tab3}, we compare the inference time required by different methods to generate a single 3D radio map. This is because fast and high-quality inference is critical for radio map estimation, particularly in high-mobility scenarios or real-time resource management applications. For fair comparison, we normalize the inference speeds of all baseline methods to the equivalent performance on an NVIDIA RTX 5090 GPU.  Owing to its end-to-end 3D structure, RadioGen3D performs one-shot inference, which thereby enables superior inference speed compared to existing RME solutions requiring layer-by-layer sequential inference or a multi-step denoising process. This advantage underscores the advantage of RadioGen3D for real-time deployment.

In addition to quantitative evaluation, visualized results are also provided in Fig. \ref{fig:Comparison at different heights} and Fig. \ref{fig:Comparison at different tx locations}, alongside the ground-truth labels for comparison. It can be observed that the visualized results obtained using the well-trained 3D U-Net clearly reproduce detailed spatial signal propagation characteristics and show high similarity to the ground-truth label. This demonstrates the capability of the proposed method for robust RME in complex 3D urban scenarios.

\subsubsection{Fine-Tuning Experiment}
We further evaluate the generalization capability of the RadioGen3D framework through fine-tuning experiments on an additional 3D radio map dataset generated via ray-tracing, termed Radio3D-RT. This also enables a direct and fair comparison with existing baselines developed on similar ray-tracing simulated datasets. Different from Radio3DMix that combines generative models with parametric fitting based on real measurements, Radio3D-RT embodies a distinct, physics-simulated data distribution. This discrepancy allows us to examine the adaptability of the 3D U-Net generator to data featuring unfamiliar propagation characteristics through limited fine-tuning. The Radio3D-RT dataset and experimental results are presented in the following.

The Radio3D-RT dataset is constructed based on $100$ building maps of scale $256\times256\times50$ $\text{m}^3$ extracted from the same Beijing urban map used for Radio3DMix. 
For each building map, $50$ 3D radio maps are generated, each corresponding to a single transmitter with random location and power, yielding a total of $5000$ data samples. Note that this is different from the Radio3DMix dataset where each radio map includes two transmitters. Moreover, ideal omni-directional antenna is employed in Radio3D-RT without considering the antenna polarization effect. These key differences in antenna modeling and transmitter configuration underscore the distribution shift between the two datasets, justifying the necessity of fine-tuning.

To align with the previous experiments, the fine-tuning process is conducted under identical configurations with sparse measurements at sampling rates of 1\% and 10\%. The transmitter-only configuration is not included as the 3D U-Net generator depends on sparse measurements to infer antenna polarization characteristics. In fact, forcing the model to adapt to an omnidirectional antenna case without sparse measurement inputs can possibly compromise its performance.
The fine-tuning process is composed of $50$ epochs. Quantitative results on the Radio3D-RT dataset before and after fine-tuning are summarized in Table \ref{tab2}.  After limited fine-tuning, the proposed model achieves consistent and competitive accuracy, outperforming existing baselines even at lower sampling rates. For example, at a 1\% sampling rate, the fine-tuned 3D U-Net generator attains an RMSE of $0.0278$ and an NMSE of $0.0199$, which is unachievable for the RadioDiff-3D baseline even at a $10\%$ sampling rate. In addition to quantitative evaluation, Fig. \ref{fig:Comparison of fine-tuning experiments at different heights} also presents visualized results after fine-tuning across multiple height slices. These visualized slices show that the fine-tuned 3D U-Net generator is capable of accurately reproducing detailed spatial signal propagation characteristics. The predictions exhibit strong structural consistency with the ground-truth labels at all heights, indicating that the model successfully transfers spatial and propagation priors learned from Radio3DMix to the ray-tracing-based dataset.

In fact, the fine-tuning method employed in this paper can also be regarded as a transfer learning strategy. Pre-training on the Radio3DMix dataset, which contains rich and diverse feature representations, allows the model to learn robust priors about 3D signal propagation, such as height-dependent channel fading and spatial correlations. Theoretically, this pre-training equips the model with substantial background knowledge, initializing its parameters at a favorable starting point for efficient adaptation to the target Radio3D-RT dataset. This significantly reduces computational costs compared to training from scratch and alleviates the issues of poor estimation accuracy and generalization, which are often associated with training deep generative models on specific, limited datasets from a random initialization.

To further validate the necessity and superiority of the proposed transfer learning strategy, we compare RadioGen3D trained from scratch on the Radio3D-RT dataset against our strategy of pre-training on Radio3D-Mix followed by fine-tuning on Radio3D-RT. For fairness, the experimental setup is rigorously controlled, with identical configurations in network architecture, loss functions, optimizer, and hyperparameters. Table \ref{tab2} presents the performance gap between the two training strategies. The fine-tuned model consistently outperforms the model trained from scratch under the sparse-and-transmitter configuration. 
Although the model trained from scratch captures the general characteristics of signal propagation, it yields inferior reconstruction fidelity for intricate spatial structures and fine-grained details. These results indicate that the pre-training on Radio3DMix provides crucial prior knowledge of physical signal propagation, which significantly enhances the generalization capability of the 3D U-Net for its adaptation to the target ray-tracing data. Such a benefit is challenging to achieve when learning directly from the target dataset with a random initialization. However, Table \ref{tab2} also indicates that the fine-tuned model leads to an inferior performance under the sparse-only configuration at low sampling rates. This is because the model, pretrained on two-transmitter scenarios, becomes dependent on either explicit transmitter information or dense measurements to adapt to single-transmitter scenarios. Consequently, this issue can be easily mitigated by generating data with a varying number of transmitters, which will be explored in future work.

In summary, the fine-tuning experiments demonstrate the effectiveness of the proposed RadioGen3D framework and validate its two key contributions, which are highlighted below.
\begin{itemize}
    \item \textbf{Data Sufficiency and Effectiveness}: According to the description Sec. \ref{sec-data-synthesis}, the proposed data synthesis approach can generate an arbitrarily large number of 3D radio map samples, significantly surpassing ray-tracing methods in terms of data generation efficiency. However, the value of the massive data volume requires validation. The fine-tuning experiments confirm that pre-training on such a large-scale synthetic dataset substantially improves estimation accuracy after adaptation. These results collectively validate the effectiveness and novelty of the proposed data synthesis and training strategy.
    \item \textbf{Strong Generalization Capability}: Despite significant differences in antenna modeling and transmitter configuration between Radio3DMix and Radio3D-RT, the model achieves high accuracy after fine-tuning.  This successful transfer of knowledge not only proves the strong generalization capability of the 3D U-Net generator but also confirms the effectiveness of the data synthesis idea proposed in Sec. \ref{sec-data-synthesis}. Specifically, by covering diverse signal propagation characteristics with multiple coefficient combinations, the synthetic data exhibits high fidelity to both simulated and real scenarios.
\end{itemize}

\section{Conclusion}\label{sec:conclusion}
In this work, we propose RadioGen3D, a general framework for 3D RME with two key contributions including the efficient synthesis of a large-scale, high-quality 3D radio map dataset Radio3DMix, and a generative model training scheme based on adversarial learning. The effectiveness and superiority of RadioGen3D are demonstrated through comprehensive experiments, showing higher estimation accuracy over existing baselines and strong generalization capability via successful fine-tuning. In summary, this work offers a novel data synthesis solution to mitigate the data scarcity issue in 3D RME, coupled with a training scheme that enables fast and accurate 3D RME. We believe it provides significant insights for future works on 3D RME in low-altitude networks, with potential applications in drone trajectory planning, network resource optimization, and digital twins for the low-altitude economy.

\bibliographystyle{IEEEtran}
\bibliography{reference.bib}

\end{document}